# Digital rock reconstruction with user-defined properties using conditional generative adversarial networks

**Qiang Zheng[1] and Dongxiao Zhang[2],***

[1]Intelligent Energy Laboratory, Frontier Research Center, Peng Cheng Laboratory, Shenzhen 518000, P. R. China

[2]School of Environmental Science and Engineering, Southern University of Science and Technology, Shenzhen 518055, P. R. China

**Key Points:**

- Conditional generative adversarial networks are proposed for digital rock reconstruction with excellent visual and geologic realism
- The conditional generative adversarial networks incorporate experimental data and user-defined properties together
- The proposed method can produce random digital rock images with user-specified rock type and first two moments

* Correspondence to: zhangdx@sustech.edu.cn

# Abstract

Uncertainty is ubiquitous with flow in subsurface rocks because of their inherent heterogeneity and lack of *in-situ* measurements. To complete uncertainty analysis in a multi-scale manner, it is a prerequisite to provide sufficient rock samples. Even though the advent of digital rock technology offers opportunities to reproduce rocks, it still cannot be utilized to provide massive samples due to its high cost, thus leading to the development of diversified mathematical methods. Among them, two-point statistics (TPS) and multi-point statistics (MPS) are commonly utilized, which feature incorporating low-order and high-order statistical information, respectively. Recently, generative adversarial networks (GANs) are becoming increasingly popular since they can reproduce training images with excellent visual and consequent geologic realism. However, standard GANs can only incorporate information from data, while leaving no interface for user-defined properties, and thus may limit the representativeness of reconstructed samples. In this study, we propose conditional GANs for digital rock reconstruction, aiming to reproduce samples not only similar to the real training data, but also satisfying user-specified properties. In fact, the proposed framework can realize the targets of MPS and TPS simultaneously by incorporating high-order information directly from rock images with the GANs scheme, while preserving low-order counterparts through conditioning. We conduct three reconstruction experiments, and the results demonstrate that rock type, rock porosity, and correlation length can be successfully conditioned to affect the reconstructed rock images. Furthermore, in contrast to existing GANs, the proposed conditioning enables learning of multiple rock types simultaneously, and thus invisibly saves computational cost.

# 1. Introduction

Due to the scarcity of observations, uncertainty exists in hydrogeological problems and should be considered when modeling flow and transport in subsurface porous media (Zhang, 2001). During the modeling process, some material properties, such as permeability, possessing huge stochasticity due to the heterogeneity of porous media, usually serve as input parameters of the physical model, and thus lead to uncertain model outputs. To evaluate the uncertainty in a systematic and multi-scale manner, some researchers focused on how randomness in the micro pore structure will affect macro permeability and eventually determine model outputs, such as pressure head (Icardi et al., 2016; Wang et al., 2018; Xue et al., 2019; Zhao et al., 2020). In terms of such uncertainty analysis workflow, the rapid and accurate reconstruction of porous media for random realizations constitutes the initial step and is critically important.

In the last few decades, advances in three-dimensional imaging techniques, e.g., X-ray computed tomography (e.g., Micro-CT and Nano-CT) (Bostanabad et al., 2018; Chen et al., 2013; Li et al., 2018), focused ion beam and scanning electron microscope (FIB-SEM) (Archie et al., 2018; Tahmasebi et al., 2015) and helium-ion-microscope (HIM) for ultra-high-resolution imaging (King Jr et al., 2015; Peng et al., 2015; Wu et al., 2017), have contributed to the development of digital rock technologies, which provide new opportunities to investigate flow and transport in rocks via numerical simulation based on digital representation of rock samples. Constrained by the high cost of three-dimensional imaging, digital rock technologies remain unsuitable for uncertainty analysis because of the lack of random rock samples, thus leading to the development of diversified mathematical methods for random reconstruction of digital rocks. These methods can be divided into the following three groups: object-based methods; process-based techniques; and pixel-based methodologies (Ji et al., 2019). Object-based methods treat pores and grains as a set of objects which are defined according to prior knowledge of the pore structure (Pyrcz and Deutsch, 2014). Such methods can be stably implemented, because they have relatively explicit objective functions, so that optimization methods, such as simulated annealing, can converge easily. However, the disadvantage is that they cannot reproduce long-range connectivity of the pore space because they only utilize low-order information. Process-based techniques, on the other hand, can produce more realistic structures than object-based methods through imitating the physical processes that form the rocks (Biswal et al., 2007; Øren and Bakke, 2002). This process is relatively time-consuming, however, and necessitates numerous calibrations.

Pixel-based methodologies work on an array of pixels in a regular grid, with the pixels representing geological properties of the rocks. Within these methods, geostatistics is the core technique that contributes to the reconstruction of rock samples, primarily including two-point statistics (TPS) and multi-point statistics (MPS) (Journel and Huijbregts, 1978; Kitanidis, 1997). One of the most commonly used TPS methods is the Joshi-Quiblier-Adler (JQA) method, which is named according to its three contributors (Adler et al., 1990; Joshi, 1974; Quiblier, 1984). The JQA method was designed to truncate the intermediate Gaussian field that satisfies a specified two-point correlation, and finally obtain a binary structure with given porosity and two-point correlation function (Jude et al., 2013). Even though two-point statistics is a very common concept in geostatistics and its related methods are easy to implement, it cannot reproduce the full characteristics of the pore structures because only low-order information is adopted. Under this circumstance, MPS was proposed to address this issue. Unlike TPS, MPS is able to extract local multi-point features by scanning a training image using a certain template, leading to the incorporation of high-order information, and thus better reproducing performance (Okabe and Blunt, 2005). Motivated by multiple applications, MPS methods are flourishing and have diversified variants, mainly comprising single normal equation simulation (SNESIM) (Strebelle, 2002), direct sampling method (Mariethoz et al., 2010), and cross-correlation-based simulation (CCSIM) (Tahmasebi and Sahimi, 2013). Although MPS features sufficiently capturing global information, it is challenged by the conditioning of local information, or equivalently honoring low-order statistics as TPS does. As a consequence, certain fundamental properties, such as volumetric proportion, may not be satisfied even though they should have been honored in the generated samples (Mariethoz and Caers, 2014).

In recent years, deep learning has achieved great success in image synthesis, primarily for facial, scenery, and medical imaging (Abdolahnejad and Liu, 2020; Choi et al., 2018; Nie et al., 2018). Due to the similar data format and task object, the success of deep learning in image synthesis inspired its application to digital rock reconstruction. The biggest advantage of deep learning is that it constitutes purely data-driven workflow with no prior information needed, and the totally learnable machine avoids complex hand-crafted feature design. Another significant benefit is that once the deep learning model is trained, prediction (generating new structures) can be accomplished within a few moments, which is precisely the bottleneck of traditional methods, such as MPS. Among deep learning methods, generative adversarial networks (GANs) (Goodfellow et al., 2014) have achieved the most popularity in digital rock reconstruction, since they can learn images in a generative and unsupervised manner, and eventually reproduce highly realistic new images that are similar to the training ones.

Concerning GANs-related applications in reconstructing digital rocks, the early work is that Mosser et al. applied GANs to learn three-dimensional images of micro structures with three rock types, i.e., bead pack, Berea sandstone and Ketton limestone, and successfully reconstructed their stochastic samples with morphological and flow characteristics maintained (Mosser et al., 2017; Mosser et al., 2018). On top of GANs, Shams et al., (2020) integrated it with auto-encoder networks to produce sandstone samples with multiscale pores, enabling GANs to predict inter-grain pores while auto-encoder networks provide GANs with intra-grain pores. Some other representative applications also exist, such as utilizing GANs to augment resolution and recover the texture of micro-CT images of rocks (Wang et al., 2019; Wang et al., 2020), and reconstructing three-dimension structures from two-dimension slices with GANs (Feng et al., 2020).

Although GANs are successfully verified to reconstruct several kinds of rocks, the information source for learning may be excessively single, i.e., only the rock images, and prior information about the rocks cannot be incorporated in the current GANs workflow. As a result, the generated samples could be too random and less representative, which may limit their potential for downstream research about pore-scale flow modeling. For instance, it is easy to produce sandstone samples with realistic porosity in previous work, but hardly possible to synthesize plentiful samples with specified porosity or other user-defined properties. In addition, current GANs are strictly developed for reconstructing rocks of a specific type, which means that it is necessary to restart GANs training when a new rock image is prepared for random reconstruction. Such kind of processing will invisibly aggravate computational burden, and thus needs to be improved.

To enable GANs to study images of different rock types simultaneously and enhance the representativeness of generated samples according to user-defined properties, we leverage the GANs in a conditioning manner, which was originally proposed by Mirza and Osindero (2014). In this study, we adopt progressively growing GANs as a basic architecture (Karras et al., 2017), which has shown excellent performance in image synthesis (Liang et al., 2020; Wang et al., 2018), and has been successfully applied in geological facies modeling (Song et al., 2021). Inspired by the work of generating MNIST digits based on class labels with conditional GANs (Mirza and Osindero, 2014), we make the rock type as one of the conditional information, aiming to generate samples with respect to a specified rock type. Moreover, the two basic statistical moments, i.e., porosity and two-point correlations, usually adopted to describe morphological features, also serve as another conditional information to produce samples with user-defined first two moments. Compared with object-based, process-based, pixel-based methods and standard GANs, the conditional GANs in our work can incorporate information not only from the data but also user-defined statistical moments (i.e., the prior information), which cannot be realized by any of the aforementioned approaches. Specifically, the proposed framework can integrate the advantages of commonly used MPS and TPS, since it is able to obtain high-order information from rock images using the GANs scheme and honor low-order moments through the conditioning manner simultaneously.

The remainder of this paper is organized as follows. Section 2 reviews the fundamental concepts and loss functions of the original GANs, and their conditional variants. Section 3 introduces the experimental data and the network architectures used in this study, and then demonstrates the experimental results with three conditioning settings. Finally, conclusions are given in section 4.

## 2. Methods

### 2.1 Generative adversarial networks

GANs were initially proposed by (Goodfellow et al., 2014) to learn an implicit representation of a probability distribution from a given training dataset. Suppose that we have a dataset **y**, which is sampled from $P_{\text{data}}$. Our goal is to build a model, named generator ($\mathcal{G}$), to generate fake samples that nearly support the real distribution $P_{\text{data}}$. The generator is parameterized by $\boldsymbol{\theta}$, and takes the random noise $\mathbf{z} \in P_{\mathbf{z}}$ as inputs to generate the fake samples $\mathcal{G}_{\boldsymbol{\theta}}(\mathbf{z})$. In order to discern the real samples from the fake samples, another model, named discriminator ($\mathcal{D}$), is adopted. The discriminator $\mathcal{D}_{\boldsymbol{\phi}}$ with parameters $\boldsymbol{\phi}$ can be viewed as a standard binary classifier, aiming to label the fake samples as zeros and the real samples as ones. Therefore, the generator and discriminator are placed in a competition, in which the generator wants to deceive the discriminator, while the discriminator attempts to avoid being deceived. When the discriminator exceeds the generator, it will also provide valuable feedback to help improve the generator and, upon reaching the Nash equilibrium point, they can both learn maximum knowledge from the training dataset.

More formally, the two-player game between $\mathcal{G}$ and $\mathcal{D}$ can be recast as a min-max optimization problem, which is defined as follows:

$$\min_{\theta} \max_{\phi} \mathbb{E}_{\mathbf{y} \sim P_{\text{data}}}[\log \mathcal{D}_{\phi}(\mathbf{y})] + \mathbb{E}_{\mathbf{z} \sim P_{\mathbf{z}}}\left[\log\left(1 - \mathcal{D}_{\phi}(\mathcal{G}_{\theta}(\mathbf{z}))\right)\right]. \quad (1)$$

In practice, $\mathcal{G}$ and $\mathcal{D}$ are trained iteratively with the gradient descent-based method, with the purpose of approximating $P_{\text{data}}$ with $P_{\mathcal{G}(\mathbf{z})}$, or equivalently, narrowing the distance between the two distributions. In the original GANs, the distance is measured by Jensen-Shannon (JS) divergence. However, JS divergence is not continuous everywhere with respect to the generator parameters, and thus cannot always supply useful gradients for the generator. Consequently, it is very difficult to train the original GANs stably. To stabilize the training performance, Wasserstein GANs (WGANs) were proposed by switching the JS divergence with Wasserstein-1 distance, which is a continuous function of the generator parameters under a mild constraint (Arjovsky et al., 2017). Specifically, the constraint is Lipschitz continuity imposed on the discriminator, and it is realized by clamping the parameters of the discriminator to a preset range (e.g., [-0.01, 0.01]) during the training process. In order to enforce the Lipschitz constraint without imprecisely clipping the parameters of the discriminator, Gulrajani et al. (2017) proposed to constrain the norm of the discriminator's output with respect to its input, leading to a WGAN with gradient penalty (WGAN-GP), whose objective function is defined as:

$$\min_{\phi} \max_{\theta} \mathbb{E}_{\mathbf{z} \sim P_{\mathbf{z}}}[\mathcal{D}_{\phi}(\mathcal{G}_{\theta}(\mathbf{z}))] - \mathbb{E}_{\mathbf{y} \sim P_{\text{data}}}[\mathcal{D}_{\phi}(\mathbf{y})] + \lambda \mathbb{E}_{\hat{\mathbf{x}} \sim P_{\hat{\mathbf{x}}}}[(\|\nabla_{\hat{\mathbf{x}}} \mathcal{D}(\hat{\mathbf{x}})\|_2 - 1)^2], \quad (2)$$

where $P_{\hat{\mathbf{x}}}$ is defined as sampling uniformly along the straight lines between pairs of points sampled from $P_{\text{data}}$ and $P_{\mathcal{G}(\mathbf{z})}$; and $\lambda$ is the penalty coefficient ($\lambda = 10$ is common in practice).

Apart from stabilizing training in the form of modifying loss function as WGAN-GP did, there is another kind of method aiming to speed up and stabilize the training process via revolutionizing its traditional scheme. Progressively growing GAN (ProGAN), proposed by Karras et al. (2017), is the most representative framework of such kind. The key idea in ProGAN lies in training both the generator and discriminator progressively, as shown in Figure 1, that is starting from low-resolution data, and progressively adding new layers to learn finer details (high-resolution data) when the training proceeds. This incremental property allows the models to firstly discover large-scale features within data distribution, and then shift attention to finer-scale details by adding new training layers, instead of having to learn the information from all scales simultaneously. Compared to traditional learning architectures, the progressive training has two main advantages, i.e., more stable training and less training time. In this work, we adopt ProGAN as a basic architecture, and also utilize the loss of WGAN-GP to further ensure stable training.

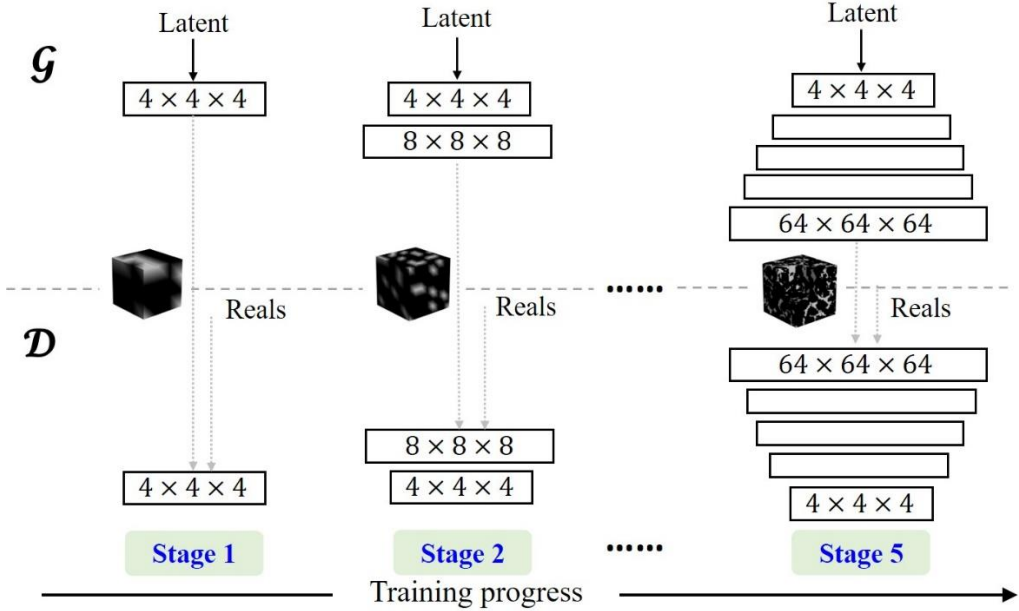

**Figure 1.** Schematic of the progressive training process. The original training data in our study are shaped as 64 × 64 × 64 voxels, and they are down-sampled to several low-resolution versions, i.e., from 32 × 32 × 32 voxels to 4 × 4 × 4 voxels with the rate as half of the previous edge length. The training procedure starts from the data with the lowest resolution, i.e., 4 × 4 × 4 voxels, and then goes to the subsequent stage to learn data with higher resolutions through adding new network layers, until the data reach the largest resolution, i.e., 64 × 64 × 64 voxels.

## 2.2 Conditional ProGAN

Although GANs are confirmed to work well in reproducing data distributions, or equivalently, generating fake samples that closely approximate the real ones, it is still worth noting that such kind of workflow is totally unconditional, which means that the training data provide the whole information that GANs can learn. However, in many cases, some conditional information needs to be considered for some specified generations, e.g., generating MNIST digits conditioned on class labels or generating face images based on gender. Likewise, in digital rock reconstruction, we aim to not only reproduce the training samples, but also allow them to satisfy some user-defined properties, e.g., porosity and two-point correlations, which can be viewed as conditional information in the generating process. Moreover, in previous work, the GANs are always trained on datasets from a single rock type and, as a consequence, the trained GANs can only generate rock samples of this kind. If one wants to obtain samples from $n$ rock types, it is necessary to separate GANs training $n$ times as previous investigations suggest (Mosser et al., 2017), which means a large amount of computational cost. Motivated by how to generate samples with different rock types using one trained GAN and enable these samples to incorporate user-defined properties, we propose a new ProGAN in a conditioning manner for digital rock reconstruction.

Conditional GANs (cGANs) were initially introduced by Mirza and Osindero (2014), aiming to generate MNIST digits conditioned on class labels. The framework of cGANs is highly similar to that of regular GANs. In the simplest form of cGANs, some extra information **c** is concatenated with the input noise, so that the generation process can condition on this information. In a similar manner, the information **c** should also be seen by the discriminator, because it will affect the distance between the real and fake distributions, which is measured by the discriminator. The original cGANs adopted the loss function in the form of JS divergence, which are deficient in training stability, as discussed in the above section. Therefore, in this work, we only borrow the conditioning manner in cGANs, and build a new loss function on top of WGAN-GP loss.

As shown in Figure 2, the generator of conditional ProGAN takes as input the three-dimensional image-like augmented noises, which are composed of random noise with known distribution and conditional labels, and they are concatenated along the channel axis. Regarding why the input noise should be formulated as image-like data, this is because the generator adopts fully convolutional architectures in order to achieve scalable generations, i.e., generate images with changeable sizes by taking noise with different shapes as input. Followed by the objective function defined in Equation (2), the loss function of the generator and the discriminator in conditional ProGAN can be written as follows:

$$\mathcal{L}_{\mathcal{G}} = -\mathbb{E}_{\mathbf{z} \sim P_{\mathbf{z}}}[\mathcal{D}_{\boldsymbol{\phi}}(\mathcal{G}_{\boldsymbol{\theta}}(\mathbf{z}, \mathbf{c}))],$$

$$\mathcal{L}_{\mathcal{D}} = \mathbb{E}_{\mathbf{z} \sim P_{\mathbf{z}}}[\mathcal{D}_{\boldsymbol{\phi}}(\mathcal{G}_{\boldsymbol{\theta}}(\mathbf{z}, \mathbf{c}))] - \mathbb{E}_{\mathbf{y} \sim P_{\text{data}}}[\mathcal{D}_{\boldsymbol{\phi}}(\mathbf{y}, \mathbf{c})] + \lambda \mathbb{E}_{\hat{\mathbf{x}} \sim P_{\hat{\mathbf{x}}}}\left[\left(\|\nabla_{\hat{\mathbf{x}}}\mathcal{D}_{\boldsymbol{\phi}}(\hat{\mathbf{x}}, \mathbf{c})\|_{2} - 1\right)^{2}\right],$$

(3)

where **c** represents the conditional labels containing rock types, porosity, and parameters of two-point correlation functions.

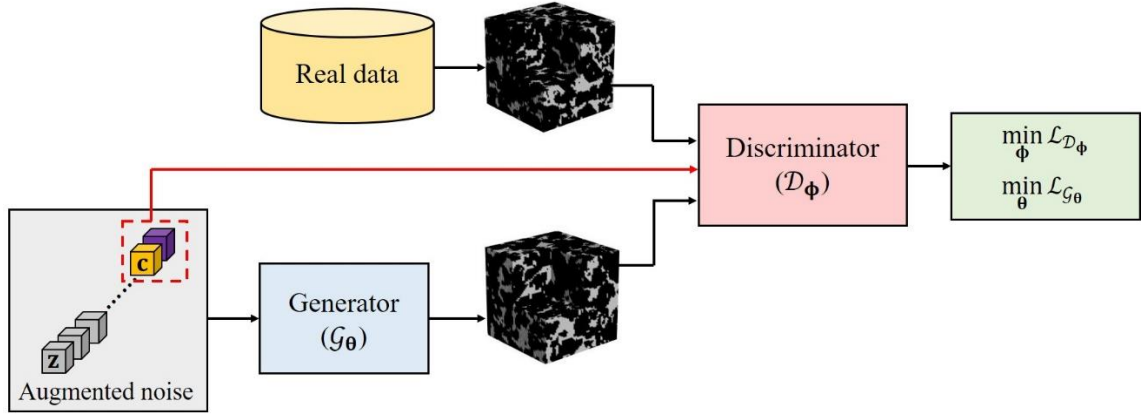

**Figure 2.** Schematic of the proposed conditional ProGAN. The input noise to the generator is augmented by conditional labels, i.e., one-hot encoded rock types, porosity, and parameters of two-point correlation functions. The labels should also serve as input of the discriminator by concatenating with the images.

## 3. Experiments

### 3.1 Experimental data

To evaluate the applicability of the proposed conditional ProGAN framework for reconstructing digital rocks, we collect five kinds of rock images, i.e., Berea sandstone, Doddington sandstone, Estaillade carbonate, Ketton carbonate and Sandy multiscale medium, from public datasets, i.e., the Digital Rocks Portal (https://www.digitalrocksportal.org/), and their basic information is listed in Table 1. The basic strategy to prepare training samples is to extract three-dimensional subvolumes from the original big sample. In the condition of limited computational resources, the size of training data cannot be very large. Meanwhile, the size of the training sample should meet the requirements of representative elementary volume (REV) to capture relatively global features (Zhang et al., 2000). Consequently, we down-sample the original rock images to $250^3$ voxels via spatial interpolations, and then we extract samples of size $64^3$ voxels with a spacing of 12 voxels, whose consequent resolutions are also listed in Table 1. Additionally, we conduct an experiment about porosity of extracted samples with respect to their edge lengths, with an aim to determine whether 64 voxels can meet the requirements of REV. As shown in Figure 3, when the edge lengths are larger than 64 voxels, the curves become much smoother than the previous parts. Even though the Sandy multiscale medium cannot get as smooth as the other four kinds of rocks when the edge length exceeds 64 voxels, it becomes much smoother than itself with smaller edge lengths. Considering that the edge length cannot be very large in order to guarantee a certain amount of extracted samples and save computational resources, we think that edge length as 64 voxels is acceptable. Finally, for each kind of rock, we can extract 4096 training samples of size $64^3$ voxels with a spacing of 12 voxels between them in the original image. To further amplify the sample size, we rotate the samples by 90° along one axis for two times, and then the total size of each training dataset can be three times of 4096, i.e., 12288.

**Table 1.** Basic information of five kinds of rock samples

| Rock type | Original size | Original resolution | Sample resolution | Reference |
|---|---|---|---|---|
| Berea sandstone | 1000×1000×1000 | 2.25 μm | 9.00 μm | (Neumann et al., 2020) |
| Doddington sandstone | 700×700×700 | 5.40 μm | 15.12 μm | (Moon and Andrew, 2019) |
| Estaillade carbonate | 650×650×650 | 3.31 μm | 8.60 μm | (Muljadi, 2015) |
| Ketton carbonate | 1000×1000×1000 | 3.00 μm | 12.00 μm | (Raeini et al., 2017) |
| Sandy multiscale medium | 512×512×512 | 3.00 μm | 6.14 μm | (Mohammadmoradi, 2017) |

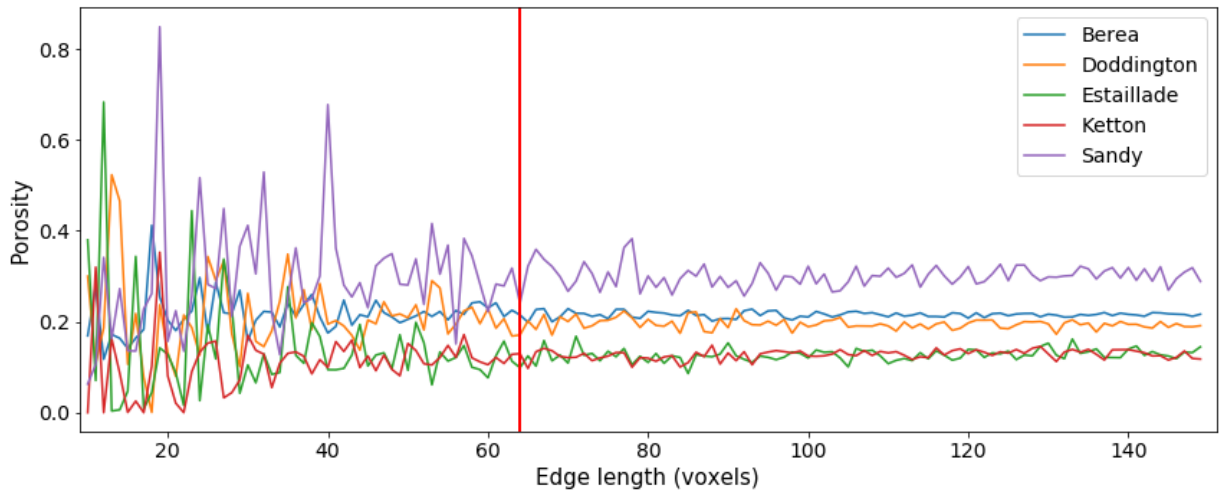

**Figure 3.** The porosity of cubic subvolumes with different edge lengths extracted from the original sample with size $250^3$ voxels. The vertical red line means that the edge length equals to 64 voxels.

For simplicity, we take the first word of names of each rock type to represent them, respectively, in this and subsequent figures.

**3.2 Network architecture**

We built the network architecture based on that of Karras et al. (2017), and made a few necessary modifications. As shown in Figure 4a, the generator network consists of five blocks, and they will be added into the network progressively once the training data enter into a higher resolution. For example, only block 1 needs to be used in stage 1 to generate fake samples with size $4 \times 4 \times 4$ voxels. When the training in stage 1 finishes and enters into stage 2, block 2 will be added to the network to produce fake samples with size $8 \times 8 \times 8$ voxels. Likewise, the subsequent stages will be conducted by introducing new blocks. To enable scalable generations, i.e., generate samples with arbitrary sizes, we replace fully-connected layers in original block 1 with convolutional layers, whose kernel size and stride are both $1 \times 1 \times 1$ voxel. Consequently, the inputs should be 5D tensors to meet the demands of convolutional operations, and here we set it as shape $N \times C \times 4 \times 4 \times 4$, in which $N$ represents batch size and $C$ means channels. In block 1, leaky ReLU activation function and pixel-wise normalization are added after the convolutional layer. To feed the discriminator single-channel data with resolution $4 \times 4 \times 4$ voxels, a convolutional layer, which adopts cubic kernel and stride with edge length as 1 voxel, is added to reduce the channels of outputs of block 1 while preserving their size unchanged. After stage 1 training, the subsequent blocks are the same in network layers, and the only difference from block 1 is that all of them adopt deconvolutional layers (with kernel size $3 \times 3 \times 3$ and strides $2 \times 2 \times 2$) to enlarge the size of feature maps by two times of the previous ones. As the same in stage 1, there is also a convolutional layer being arranged after each block to decrease the channels of their outputs. The input $\mathbf{z}^*$ to the generator is an augmented vector, which is obtained by the random noises $\mathbf{z}$ concatenated by the conditional labels $\mathbf{c}$ along the channel axis, i.e., $\mathbf{z}^* = \mathbf{z} \oplus \mathbf{c}$, where $\oplus$ represents the concatenation operation.

The discriminator network (see Figure 4b) is almost symmetric to the generator network with several exceptions. Likewise, the discriminator has the same number of blocks as the generator since they should undergo the paired training stages, which means that the outputs of the generator and the inputs of the discriminator must have the same resolution in each training stage. The input $\mathbf{Y}^*$ to the discriminator is also an augmented vector like that to the generator, and the conditional labels **c** should be reformulated as 5D tensors with the same width as training images **Y** so that they can be concatenated along the channel axis, i.e., $\mathbf{Y}^* = \mathbf{Y} \oplus \mathbf{c}$. In each stage, the training data will firstly go through a convolutional layer, whose cubic kernel and stride have edge length as one voxel, to enlarge channels while maintaining the size unchanged. After that, the outputs will undergo several blocks, which all contain a convolutional layer (with kernel size $3 \times 3 \times 3$ and strides $2 \times 2 \times 2$) and leaky ReLU activation. Finally, the outputs of block 1 will be transformed to one-dimensional scores by a fully-connected layer. It is worth noting that, when entering new training stages, the newly added blocks to generator and discriminator will fade in smoothly, with an aim to avoid sudden shocks to those already well-trained and smaller-resolution blocks. For additional details about the network architecture and training execution of ProGAN, one can refer to Karras et al. (2017).

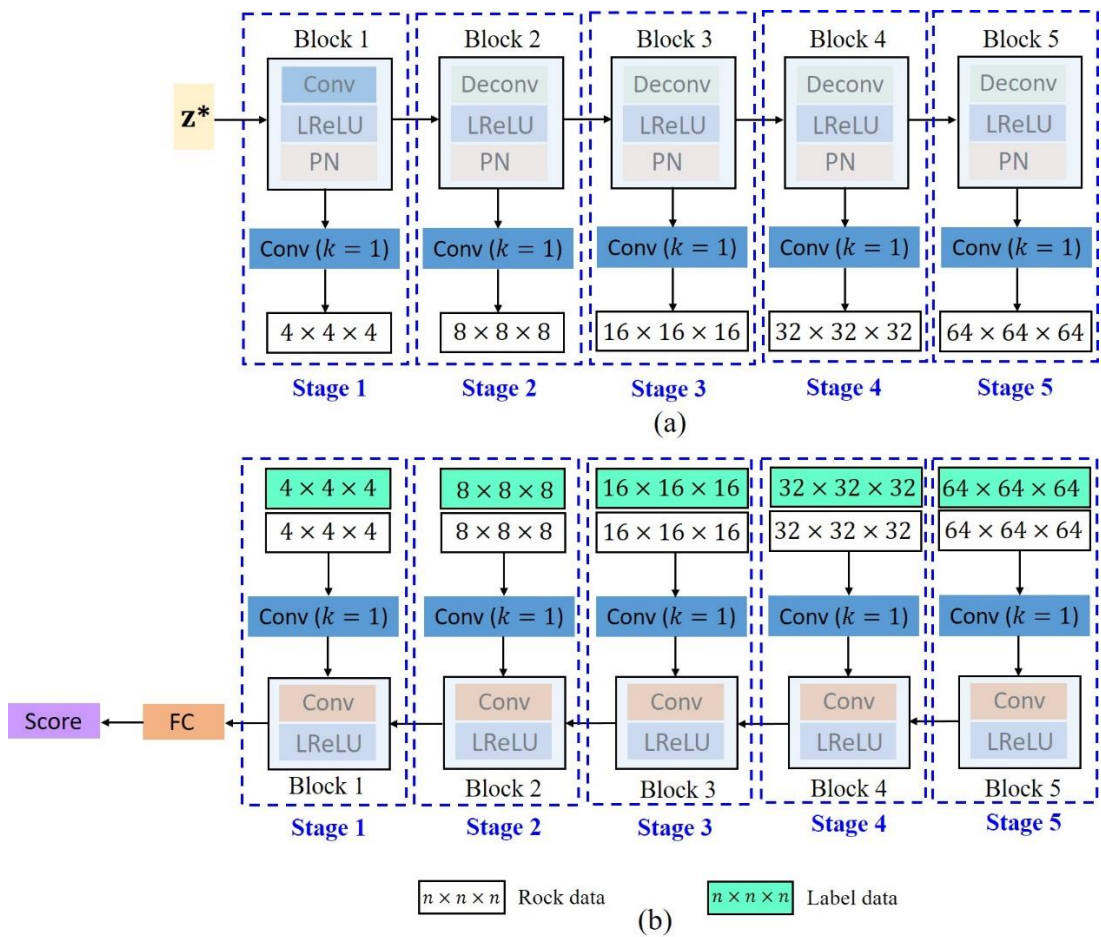

**Figure 4.** The network architecture of (a) generator and (b) discriminator. Due to the progressive training scheme, the training process is split into five stages, and they are paired for the generator

and the discriminator, which means that the output of the generator should share the same size as the input of the discriminator in each training stage. When entering new stages, new blocks will be added progressively to handle higher-resolution data. The input tensor $\mathbf{z}^*$ for the generator, augmented by random noise and labels, firstly goes through a convolutional layer (Conv), a Leaky ReLU activation layer (LReLU), and a pixel-wise normalization layer (PN) in block 1. The subsequent blocks are the same as block 1 except for the deconvolutional layer (Deconv), which is used for enlarging the size of feature maps. In each stage, the output of the final block will go through a Conv layer with kernel and stride size as 1 to reduce data channels and obtain fake data with the same shape as training data. The discriminator takes multi-resolution data and labels as input, which share the same size and concatenate along the channel axis. The input data firstly go through a Conv layer with kernel and stride size as 1 to increase channels, and then pass some blocks, which are all composed of a Conv layer (with cubic kernel size as 3 and strides as 2) and a LReLU activation layer. Finally, a fully-connected layer (FC) is used to transform the outputs of block 1 to one-dimensional scores.

## 3.3 Experimental results

### 3.3.1 Reconstruction conditioned on rock type

Based on the training samples of five rock types prepared above, we train the conditional ProGAN networks using the Adam optimizer (Kingma and Ba, 2014). The learning rate ($lr$) is set based on the current training stage or data resolution, and here we set $lr = 5\mathrm{e} - 3$ when the resolution is less than or equal to $16^3$ voxels, $lr = 3.5\mathrm{e} - 3$ for resolution as $32^3$ voxels, and $lr = 2.5\mathrm{e} - 3$ for resolution as $64^3$ voxels. In this section, we consider the rock type as conditional labels, aiming to study several kinds of rock data simultaneously with one model rather than wasting time to build several models. The conditional rock type are discretized labels, and thus should be one-hot encoded so as to concatenate with random noise. As mentioned in Section 3.2, we set the inputs of generator ($\mathbf{z}^*$) as shape $N \times C \times 4 \times 4 \times 4$, and thus the random noise $\mathbf{z}$ sampled from standard Gaussian distribution $P_\mathbf{z}$ has shape $N \times 1 \times 4 \times 4 \times 4$. To concatenate with image-like noise along the channel axis, the original label with size $N \times 5$ should be reshaped and repeated as a tensor with size $N \times 5 \times 4 \times 4 \times 4$. Likewise, when feeding the discriminator, the labels should also be reshaped and repeated so as to concatenate with multi-resolution rock data. Therefore, in this case, the channel $C$ of inputs to the generator and the discriminator is 6.

The batch size $N$ is chosen as 32 for each training stage in this work. In each stage, we set 320,000 iterations of alternative training when data resolution is less than or equal to $16^3$ voxels, and 640,000 iterations for all larger resolutions. We train the networks for 2,880,000 iterations in total, which requires approximately 23 hours running time on four GPUs (Tesla V100).

With the trained model, we can produce new realizations by feeding the generator with new random noise sampled from $P_z$ and the specified rock type. Here, in its simplest form, we assign the one-hot code [1, 0, 0, 0, 0] as Berea sandstone, [0, 1, 0, 0, 0] as Doddington sandstone, and so on in a similar fashion for other rocks. Since we utilized a fully convolutional architecture to build a scalable generator, we can produce new samples with different sizes by feeding the generator with noise of different shapes. During the model training, the input $z^*$ is shaped as cubic image-like data with edge length as 4 voxels and the output is also cubic data with side length as 64 voxels. To make the generated cubic samples become larger, we increase edge length of $z^*$ as 6, 8 and 10 voxels to obtain the corresponding outputs with edge length as 96, 128, and 160 voxels. We randomly select one sample for each kind of rock to visualize the reconstruction performance. As shown in Figure 5, the left column represents training samples, while the others are generated samples with different sizes. Firstly, the reconstructed rocks of different kinds are distinguishable, but visually similar to their own kinds, which means that the conditioning of rock type does work. Secondly, the scalable generator can effectively produce samples with larger sizes than those of training ones without losing morphological realism. It is worth noting that the reconstruction of multiscale medium is difficult for normal GANs, and therefore Shams et al. (2020) proposed a method by integrating GANs and auto-encoder to specifically produce multiscale medium. In this work, owing to the great power of the progressive training scheme, the multiscale features can be naturally incorporated into the model, and thus there is no need to design a specific model for multiscale medium.

Apart from visual realism, we need to quantitatively evaluate the reconstruction performance by computing statistical similarities between training samples and generated ones. In the original ProGAN work, Karras et al. (2017) asserted that a good generator should produce samples with local image structures similar to the training set over all scales, and they proposed multi-scale sliced Wasserstein distance (SWD) to evaluate the multi-scale similarities. The multi-scale SWD metric is calculated based on the local image patches drawn from Laplacian pyramid representations of generated and training images, starting at a low resolution of $16^3$ voxels and doubling it until reaching the full resolution. Here, we sample 4,000 training and generated images, and randomly extract 32 7 × 7-pixel slice patches from the Laplacian pyramid representation of each image at each resolution level, to prepare 128,000 patches from the training and generated dataset at each level. Since each saved model during the training process can be utilized to calculate multi-scale SWD, we can obtain the changes of SWD with respect to the iteration steps at each level. In this work, we average the multi-scale SWD over different levels to acquire a mean value to evaluate the distance between two distributions. Meanwhile, we also calculate the average SWD at the highest level of real samples as a benchmark, i.e., similarities between real patches, and their values are $7.23 \times 10^3$, $7.41 \times 10^3$, $7.22 \times 10^3$, $7.31 \times 10^3$, and $7.50 \times 10^3$ for Berea sandstone, Doddington sandstone, Estaillade carbonate, Ketton carbonate, and Sandy multiscale medium, respectively. As shown in Figure 6, the average SWD curves of generated samples from five rocks can converge to a relatively small value, and are extremely close to the above benchmark values, which means that when we stop training, the generator can produce very realistic samples.

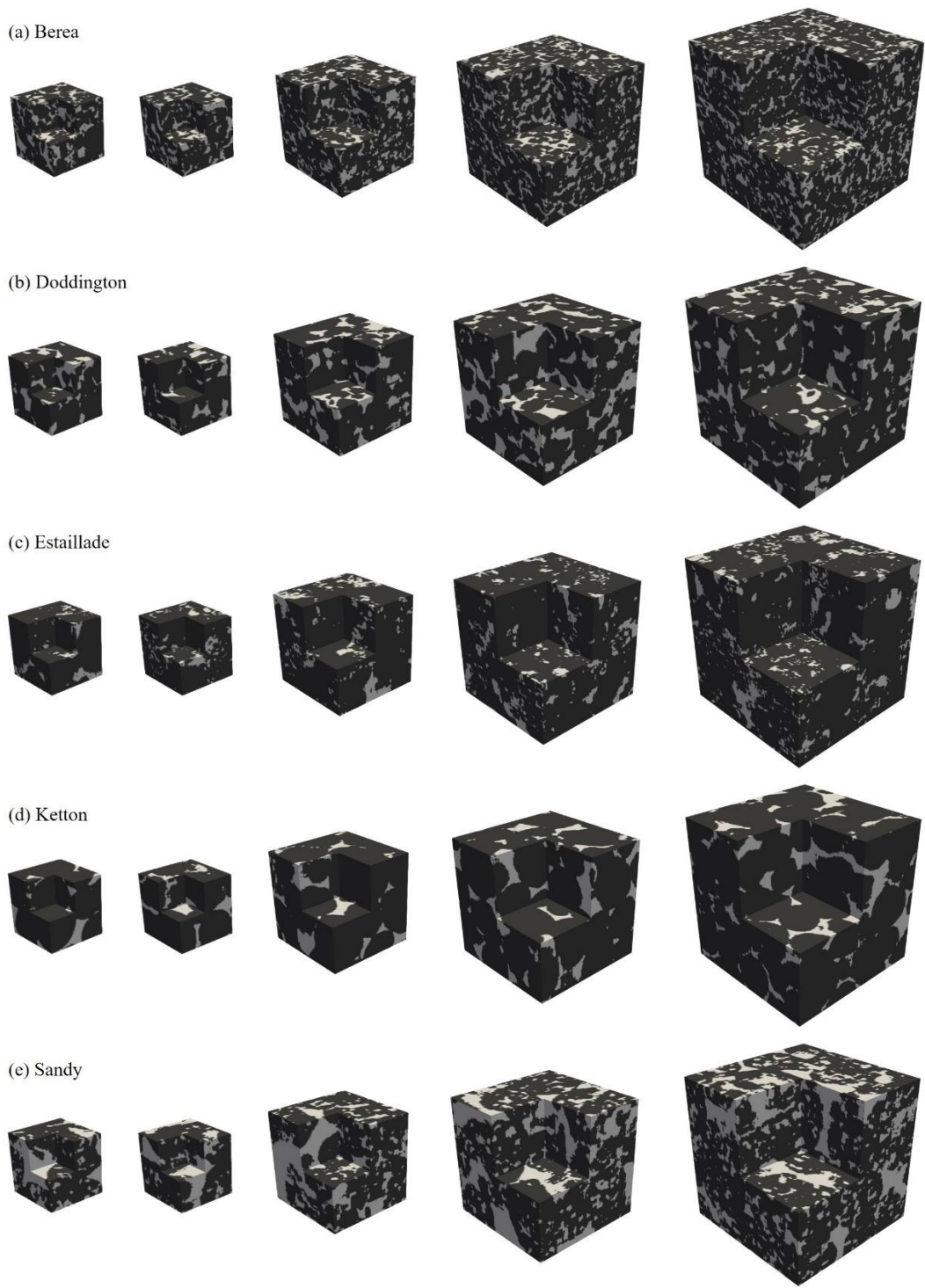

**Figure 5.** Training samples (the first column in each subfigure) and synthetic samples (the subsequent four columns in each subfigure) with different sizes of five rock types: (a) Berea

sandstone; (b) Doddington sandstone; (c) Estaillade carbonate; (d) Ketton carbonate; and (e) Sandy multiscale medium.

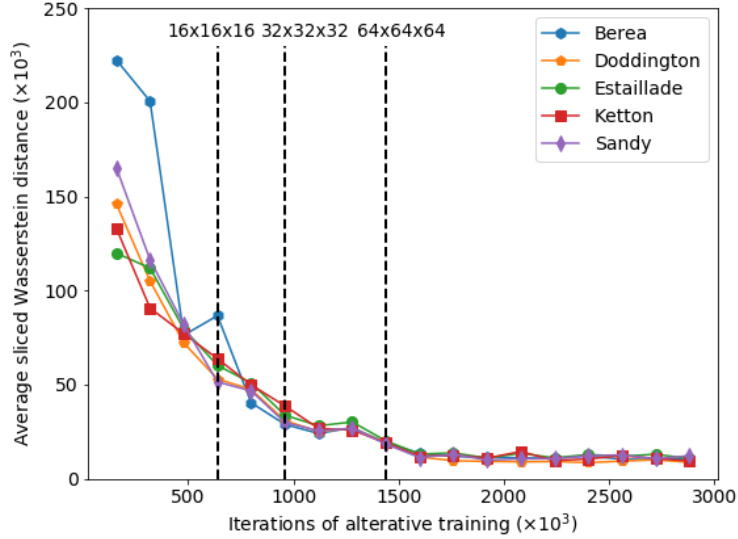

**Figure 6.** Average sliced Wasserstein distance (SWD) of five rock types during the training process.

In addition to using SWD to assess the generative performance, we also conduct geostatistical analysis to further verify the morphological realism of reconstructed samples. The rock media in this experiment can be defined as a binary field $F(\mathbf{x})$ as follows:

$$F(\mathbf{x}) = \begin{cases} 1 & \mathbf{x} \in \Omega_{\text{pore}} \\ 0 & \mathbf{x} \in \Omega_{\text{solid}} \end{cases}, \tag{4}$$

where $\mathbf{x}$ represents any point in the image of porous media; and $\Omega_{\text{pore}}$ and $\Omega_{\text{solid}}$ are the space occupied by the pores and solid grains, respectively. In order to characterize the structures of rocks, the first- and second-order moments of $F(\mathbf{x})$ defined as follows can be utilized:

$$\phi_F = \overline{F(\mathbf{x})}, \tag{5}$$

$$R_F(\mathbf{x}, \mathbf{x} + \mathbf{r}) = \frac{\overline{[\phi_F - F(\mathbf{x})] \cdot [\phi_F - F(\mathbf{x} + \mathbf{r})]}}{\phi_F - \phi_F^2}, \tag{6}$$

where $\phi_F$ represents the porosity; and $R_F(\mathbf{x}, \mathbf{x} + \mathbf{r})$, termed the normalized two-point correlation function, represents the probability that two points $\mathbf{x}$ and $\mathbf{x} + \mathbf{r}$, separated by lag vector $\mathbf{r}$, are located in the pore phase $\Omega_{\text{pore}}$. In addition, we also calculate specific surface area to measure morphological similarities, which is expressed as:

$$S_a = \frac{1}{V} \int dS, \tag{7}$$

where integration occurs at the solid-pore interface $S$; and $V$ is bulk volume.

In order to calculate the above three metrics, we randomly select 200 samples from the training dataset with respect to each rock type, and simultaneously generate new realizations with the same amount. As shown in Figure 7, the porosity ranges of generated samples of size $64^3$ voxels agree well with those of training ones for all rock types. When enlarging the size of generated samples, their porosity ranges become narrower while keeping median values close to the real ones. This can be easily understood from the perspective of stochastic process, in which the random field with larger scale is more closely approaching ergodicity, and thus its statistical properties, such as porosity, should be more stable than that with a smaller size (Papoulis and Pillai, 2002; Zhang, 2001). The normalized two-point correlation curves in three dimensions for each kind of rock are presented in Figure 8. Obviously, both the means and the ranges of the two-point correlation functions of generated samples can match well with those of the original training ones. When samples become larger, the two-point correlation functions have narrower ranges and longer tails, and keep the mean curves tightly close to the real ones. This means that the second-order structure of rock media can be maintained when we produce larger samples than the training ones. The comparisons of specific surface area are shown in Figure 9. Likewise, this metric also matches well between the training samples and the synthetic ones, and approaches median values when the sample size becomes larger.

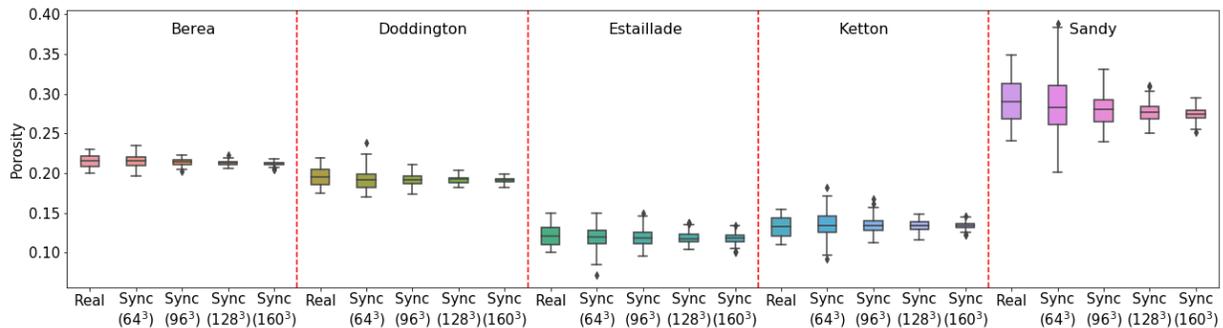

**Figure 7.** Comparisons of the porosity of real training samples and synthetic (Sync) samples with different sizes of five rock types.

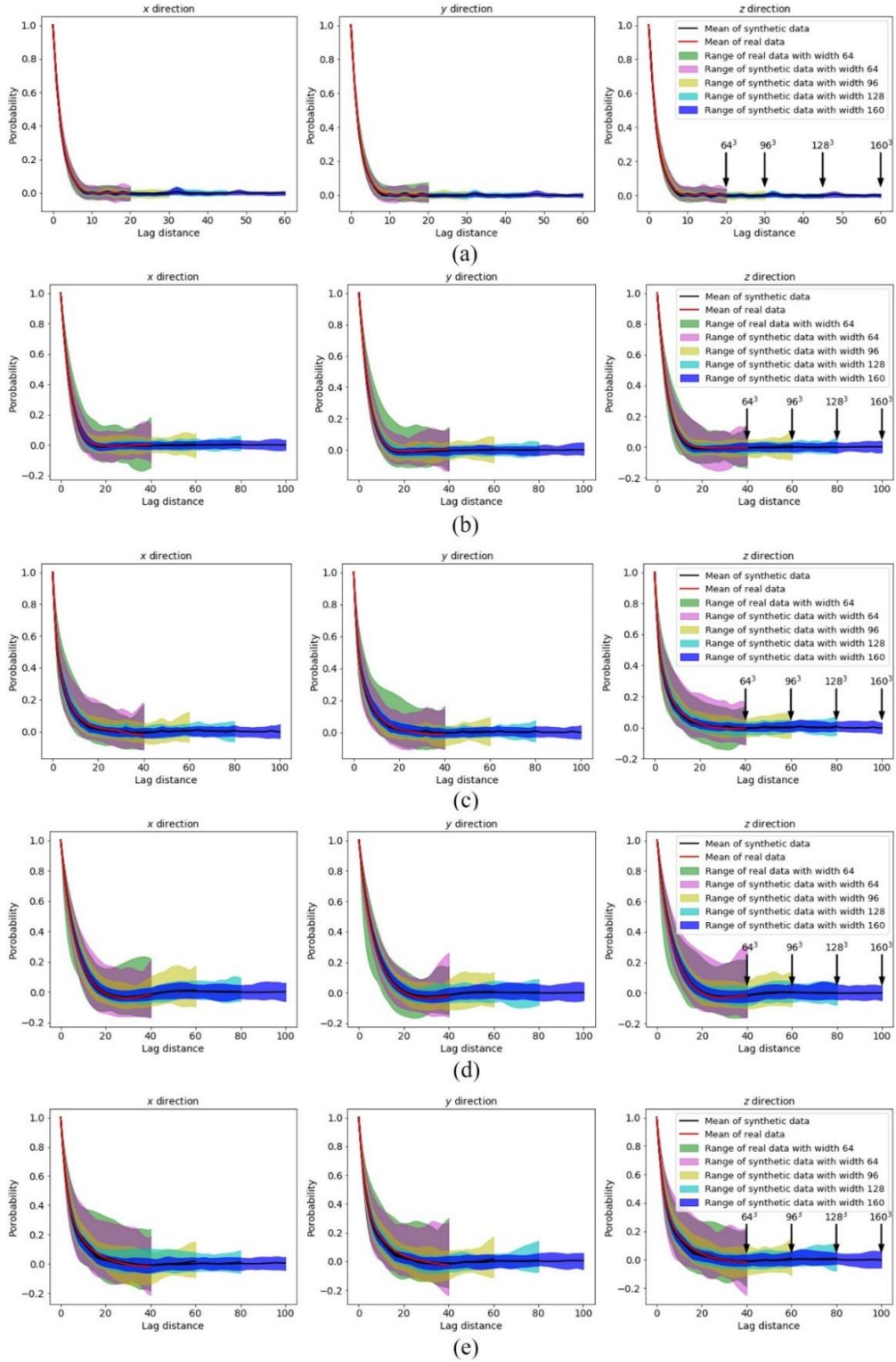

**Figure 8.** Comparisons of the normalized two-point correlation function of training samples and synthetic samples with different sizes of five rock types: (a) Berea sandstone; (b) Doddington sandstone; (c) Estaillade carbonate; (d) Ketton carbonate; and (e) Sandy multiscale medium.

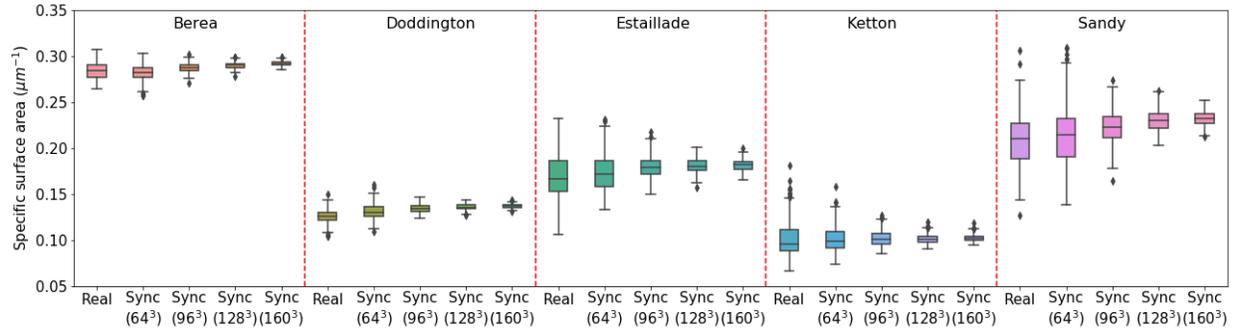

**Figure 9.** Comparisons of the specific surface area of training samples and synthetic samples with different sizes of five rock types.

To evaluate the physical accuracy of reconstructed samples, we calculate their absolute permeability using the single-phase Lattice Boltzmann method (Eshghinejadfard et al., 2016). In this case, we ignore the anisotropy of each kind of rock and plot the results in Figure 10. It is obvious that the permeability of generated samples matches well with that of training ones, and also gets closer to median values when enlarging sample size. Furthermore, with sample size increasing, the abnormal values of permeability become less and less. This means that the larger are the samples, the more probable it is for them to achieve realistic and stable morphological and physical properties, which is crucial for downstream research of pore-scale flow based on the reconstructed samples.

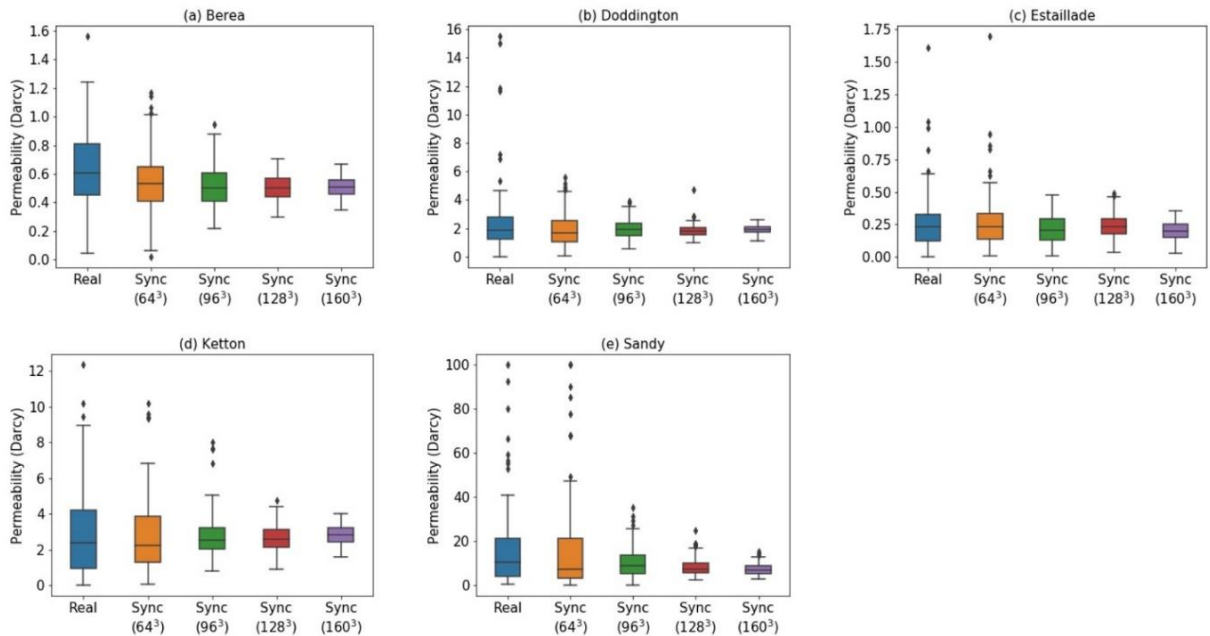

**Figure 10.** Comparisons of the permeability of real training samples and synthetic samples with different sizes of five rock types.

### 3.3.2 Reconstruction conditioned on porosity

After validating that the conditioning manner worked successfully on rock type, we continue to consider the first-order moment, i.e., the porosity, as another kind of conditional label with an aim to generate random realizations with a specified porosity. During the model training, the scalar porosity labels should be reshaped and repeated as the same dimension with inputs for the generator and the discriminator, respectively, just as it was done in the last section for rock type. When training the discriminator, i.e., minimizing $\mathcal{L}_\mathcal{D}$ in Equation (3), the conditional porosity should be the corresponding values of the training samples. Meanwhile, when training the generator, i.e., minimizing $\mathcal{L}_\mathcal{G}$ in Equation (3), we hope that the generator can see reasonable porosities with equal possibilities, and thus we extract the maximum and minimum values of porosity from the training dataset of each kind of rock, and then uniformly sample from that range to feed the generator. In this case, we select three rock types, i.e., Doddington sandstone, Estaillade carbonate and Sandy multiscale medium, to test the performance of porosity conditioning.

With the trained model, we can produce random rock realizations with specified porosity by feeding the generator new random noise $\mathbf{z} \sim P_\mathbf{z}$ after concatenating it with a given porosity label. In this case, we set different porosity targets with respect to different rock types, i.e., $\phi_{\text{target}} = 0.21$ for Doddington sandstone, $\phi_{\text{target}} = 0.10$ for Estaillade carbonate, and $\phi_{\text{target}} = 0.22$ for Sandy multiscale medium. To validate the performance of conditioning on porosity targets, we produce 200 samples with different sizes for each rock type, and also randomly select training samples with the same amount, to calculate their porosities. It can be seen from Figure 11 that the porosities of the generated samples with different sizes all exhibit a global offset from those of the training ones to the preset targets, and their median values are rather close to the targets. Furthermore, when the samples get larger, their porosities will approach targets in a more determinate manner. In other words, the large samples are more reliable when we want to reconstruct rock samples with a specified porosity.

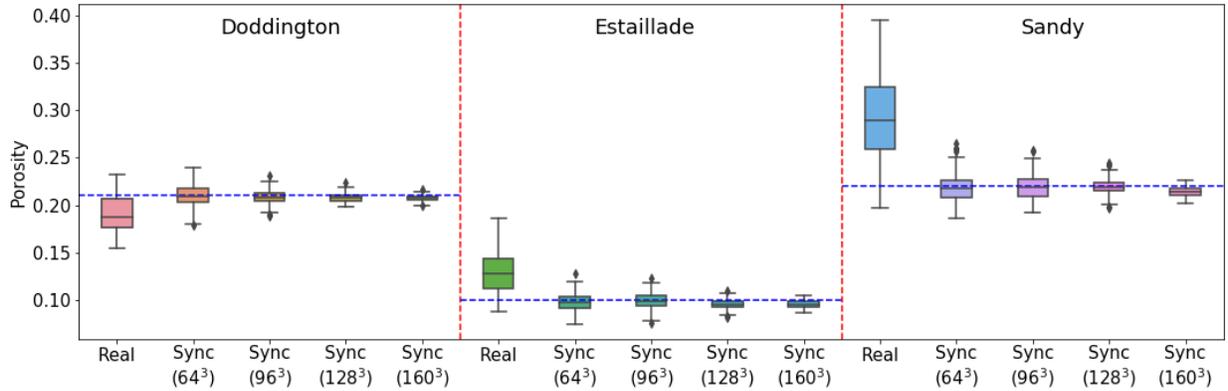

**Figure 11.** Porosity of the training samples and generated samples with different sizes for three rock types. The blue dashed line represents the preset porosity targets.

To further vividly elucidate the performance of porosity conditioning, we fix the input noise and only change the porosity for each rock type when feeding the generator to produce samples with size $160^3$ voxels. As shown in Figure 12, almost all pores become larger when gradually increasing the conditional porosity. In contrast, the patterns of the pore structure remain almost unchanged from left to right in the figure due to the fixed noise. This phenomenon reveals that the conditional information could be disentangled from the random noise, and they respectively control different aspects in the reconstructed samples, with pore structure generally encoded in the input noise and pore size controlled by the conditional label. Among the three rock types, Estaillade carbonate is visually the most heterogeneous, which we think is implicitly understood by the trained model, because the originally large pores become larger when increasing the conditional porosity, while allowing the relatively small pores to be less changed to maintain heterogeneity. It is also interesting to find that the changes in Sandy multiscale medium mainly lie in the inter-grain pores, whose sizes grow dramatically when increasing the global porosity, while keeping the inner-grain pores less changed.

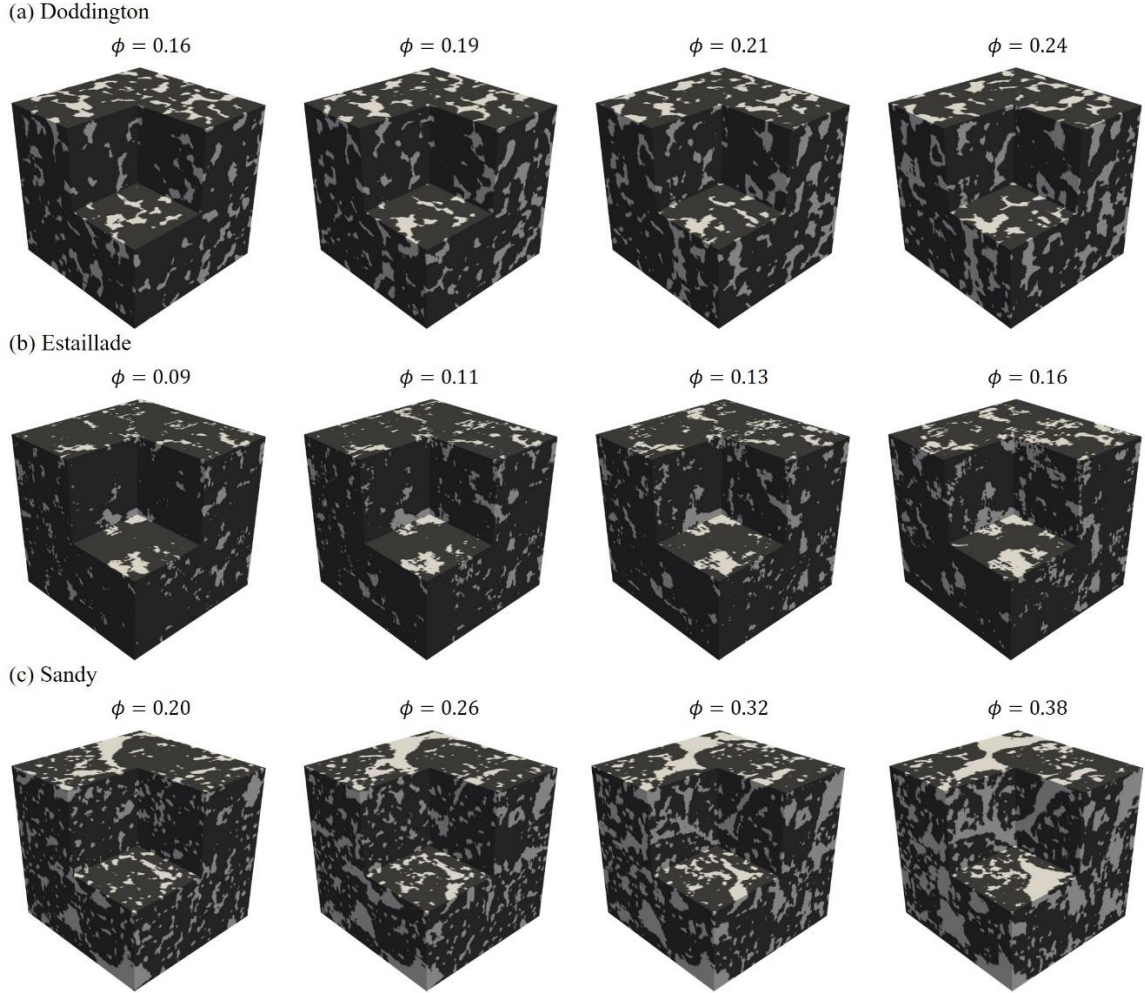

**Figure 12.** Synthetic samples of varying porosity with size $160^3$ voxels. The three rows represent Doddington sandstone, Estaillade carbonate, and Sandy multiscale medium, respectively.

Apart from evaluating conditioning performance in terms of porosity, we also investigate the effects of changing porosity on physical property. Based on the same samples used in Figure 11, we calculate the corresponding absolute permeability via the Lattice Boltzmann method. As shown in Figure 13, the permeability of the generated samples of Doddington exhibits a slight increase compared to that of training samples, since its porosity target is located in the upper quartile and larger than major samples. Estaillade presents a similar, but opposite, trend. Actually, the permeability of these two rocks did not respond sensitively to the porosity changes. In contrast, the permeability of Sandy presents remarkable decreases when setting the porosity target within the lower quartile. Regarding the reason for this, we guess that the inter-grain pores, which affect discharge capacity to a large extent, shrink dramatically when decreasing the porosity, as denoted in Figure 12c, and mainly contribute to the decrease of permeability. On the other hand, in Doddington or Estaillade, even though the pore sizes are changing, the pore throats may change a little, which have direct effects on rock permeability. Therefore, we can find that porosity conditioning can actually work through the proposed method. However, whether the changing porosity will affect permeability effectively is different among rock types, which requires further and systematic investigations but goes beyond the scope of this article.

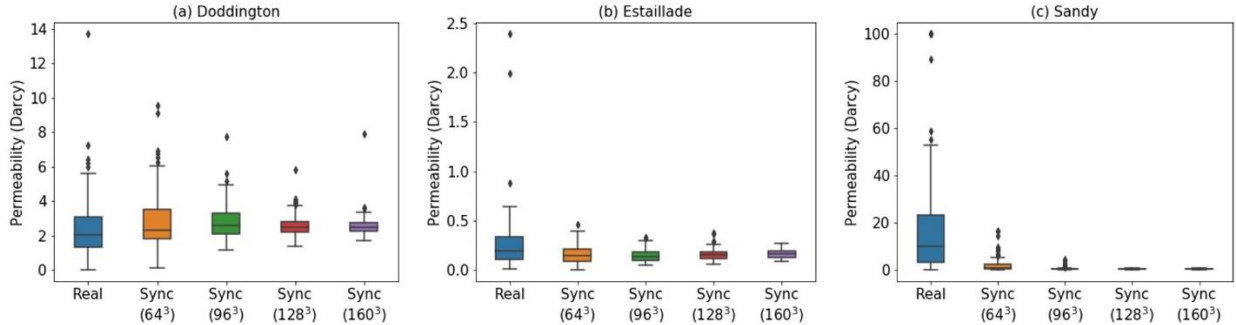

**Figure 13.** Permeability of the training samples and generated samples with different sizes for three rock types. The generated samples are the same as those in Figure 11, which adopt different preset porosity as targets.

**3.3.3 Reconstruction conditioned on correlation length**

In addition to porosity, we also test the conditioning performance of the second-order moment, i.e., the two-point correlations. Considering that the two-point correlation function is relatively hard to be conditioned on, we instead leverage a more underlying variable, i.e., the correlation length with respect to each correlation function. The normalized two-point correlation curves shown in Figure 8 are close to the exponential model, defined as follows:

$$R(r) = \exp(-r/\lambda), \qquad (8)$$

where $r$ is the lag distance; and $\lambda$ is the correlation length. They both use voxels as units in this case. The non-linear least square method can be utilized to fit $R(r)$ to the two-point correlation curves, and eventually obtain $\lambda$ corresponding to each sample. In practice, we can make use of the Python library, named *scipy.optimize*, to realize it very conveniently.

In this case, we select Berea sandstone, Ketton carbonate, and Sandy multiscale medium to test the conditioning performance. We ignore anisotropy of correlations, since we found that it is not obvious in the datasets of selected rock types. Therefore, the conditional label $\lambda$ is still a scalar value, and should concatenate with inputs for the generator and the discriminator as done for the porosity in the last section. In the same manner as the last section, when feeding the discriminator, the conditional $\lambda$ should corresponds to the training data; while feeding the generator, $\lambda$ is uniformly sampled from the range that is determined by the extremum of real labels.

After training, we can use the learned model to generate realizations by assigning a specific correlation length. In this case, we set the target correlation lengths as $\lambda_{\text{target}} = 2.4$ voxels for Berea sandstone, $\lambda_{\text{target}} = 7.0$ voxels for Ketton carbonate, and $\lambda_{\text{target}} = 3.5$ voxels for Sandy multiscale medium. We generate 200 samples for each rock type with different sizes, and randomly select the training samples with the same amount to demonstrate the performance of correlation length conditioning. As shown in Figure 14, the correlation length of generated samples with different sizes are approaching the preset targets, and their ranges become narrower with increasing size, as we found for the porosity in the previous section. Meanwhile, we plot the correlation functions of training samples and synthetic samples with size $160^3$ voxels in Figure 15. We can find that the mean curves of the generated samples get very close to the targets, and this is especially obvious for Ketton and Sandy samples.

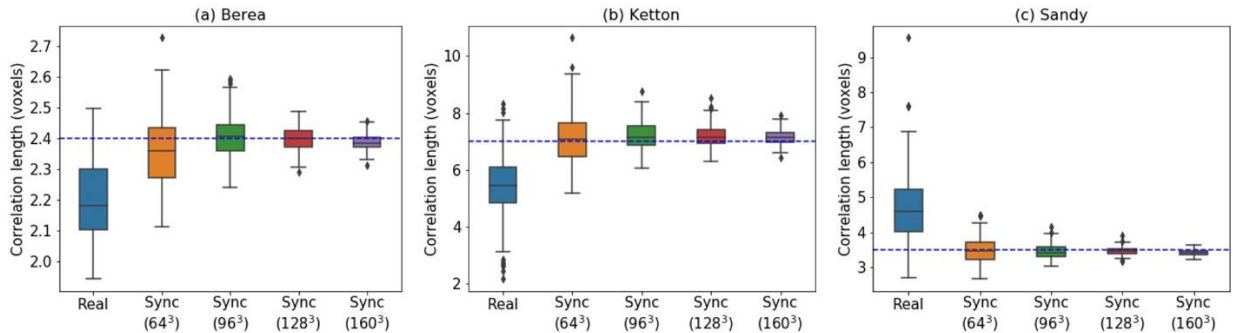

**Figure 14.** Isotropic correlation lengths (voxels) of training samples and generated samples with different sizes of three rock types. The blue dashed lines represent the respective preset targets for each rock type.

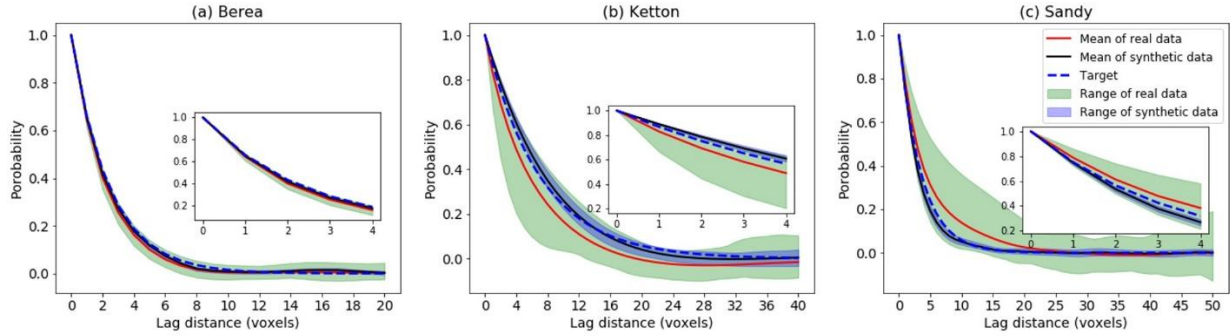

**Figure 15.** Comparisons of the normalized two-point correlation function of real and synthetic samples of three rock types. The size of synthetic samples is $160^3$ voxels.

To further clearly show the conditioning effects of correlation length, we fix the input noise and increase the correlation length gradually to produce samples with size $96^3$ voxels. Here, we choose $96^3$ since the relatively smaller size is beneficial to visualize the local changes. Theoretically, the increase of correlation length will contribute to the connection of pores. As shown in Figure 16, the Ketton sample presents the most visible changes, with pores becoming more connected from left to right in the figure. Similar to the porosity conditioning, the changes of the Sandy sample mostly lie in inter-grain pores, which become larger and thus have better connectivity when increasing correlation length. In contrast, even though the statistical metrics reveal good performance (as shown in Figure 14 and 15), little obvious variations can be found in the Berea sample, which may be partly attributed to its rather small ranges of correlation length, i.e., approximately from 2.0 to 2.5.

Furthermore, as we did in the last section, we also evaluate the effects of changing correlation length on permeability. Firstly, we calculate the absolute permeability of the plotted samples in Figure 16 to prove that correlation length does change. As shown in Table 2, the permeability of three rocks generally increases from sample 1 to sample 4 when increasing correlation length, which means that their correlation lengths did change even though the plotted samples may not reveal it very obviously. In addition, we also calculate the permeability of samples with different sizes, i.e., the samples used in Figure 14, to investigate its statistical trend. As shown in Figure 17, with preset targets larger than those of most samples, the permeability of Berea and Ketton samples exhibit a growing trend compared to training samples. In contrast, the permeability of Sandy samples decreases remarkably due to a relatively lower preset target. Therefore, it can be validated that the correlation length has a statistically positive correlation with permeability.

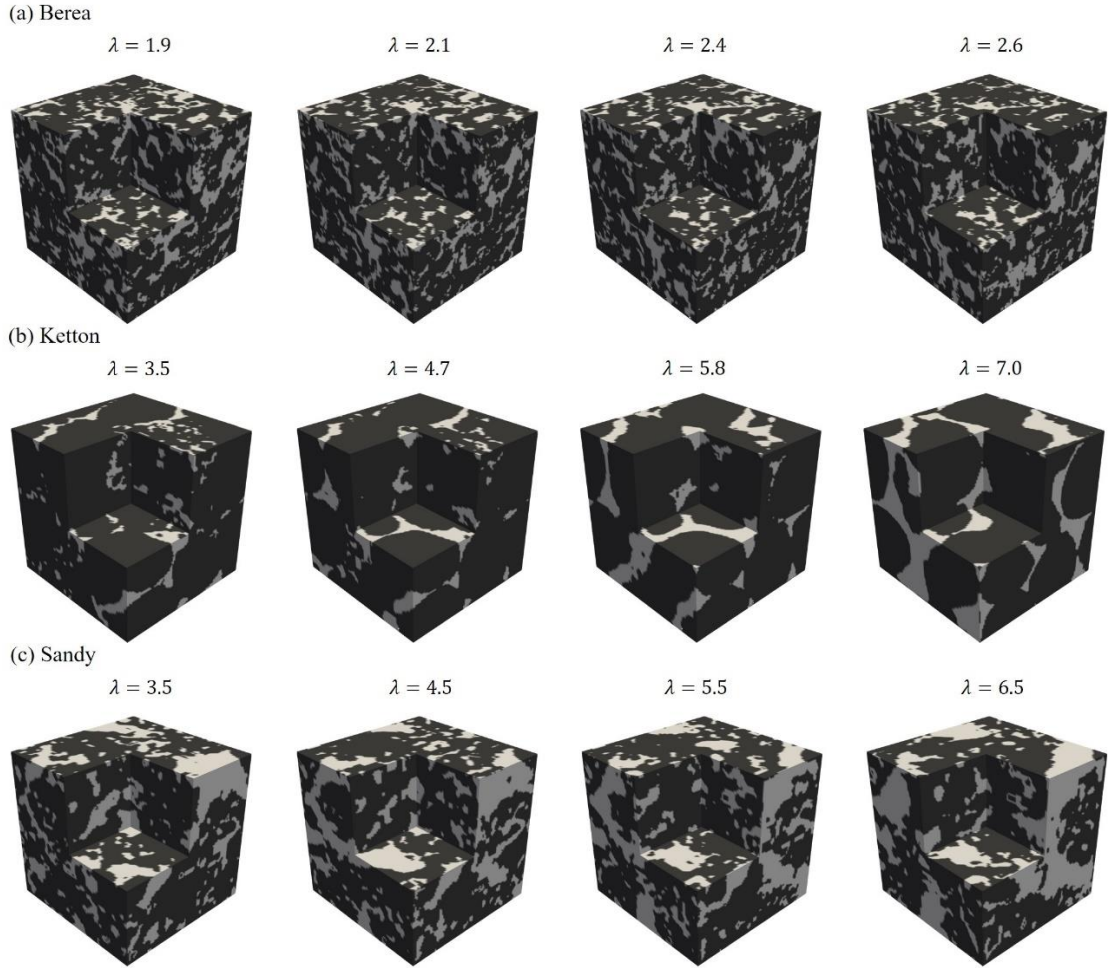

**Figure 16.** Synthetic samples with size $96^3$ voxels when only varying correlation lengths $\lambda$. The three rows represent Berea sandstone, Ketton carbonate, and Sandy multiscale medium, respectively.

**Table 2.** The permeability (Darcy) of samples in Figure 16. The correlation length increase from sample 1 to sample 4.

| Rock type | Sample 1 | Sample 2 | Sample 3 | Sample 4 |
|:---:|:---:|:---:|:---:|:---:|
| Berea | 0.56 | 0.62 | 1.08 | 1.16 |
| Ketton | 0.04 | 0.35 | 1.58 | 8.44 |
| Sandy | 0.63 | 6.04 | 9.32 | 22.03 |

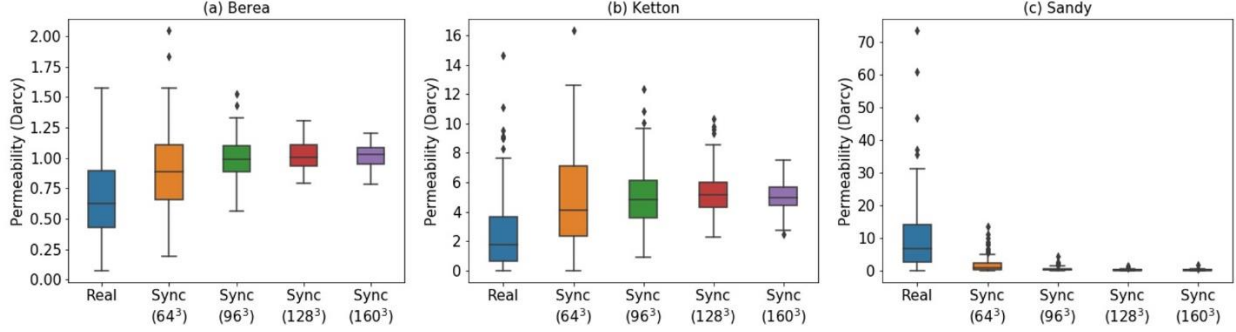

**Figure 17.** Permeability of the training samples and generated samples with different sizes for three rock types. The synthetic samples are generated by using the correlation length target set in Figure 14.

Among the rock samples listed in Table 1, we find that Doddington sandstone is relatively more anisotropic than the others, and thus we select it to further test the conditioning performance on anisotropic correlation length. Based on the initially extracted samples, i.e., 4,096 samples as stated in Section 3.1, we compute the correlation lengths of three directions according to Equation (8), and then calculate the standard deviations of correlation lengths among three directions of each sample, and rank them. We select the former 2,500 samples with relatively large standard deviations, aiming to eliminate the samples with no obvious anisotropy. Finally, we rotate the selected samples for two times as was done in Section 3.1 to obtain the eventual training dataset with size 7,500.

Upon finishing the training, we use the trained model to generate 200 samples with different sizes by setting the correlation length target as $\lambda_x = 5.0$ voxels, $\lambda_y = 3.5$ voxels, and $\lambda_z = 3.8$ voxels. As shown in Figure 18, the conditioning performance is excellent for each direction, and the samples with larger size also show better matching with corresponding targets. To demonstrate the anisotropic conditioning more clearly, we produce samples by using a fixed input noise and a gradually changing correlation length in one direction. We fix $\lambda_y = 3.5$ voxels and $\lambda_z = 3.8$ voxels, and increase $\lambda_x$ from 3.5 voxels to 6.0 voxels with roughly equal intervals (i.e., 3.50, 4.13, 4.75, 5.38, 6.00), and randomly select a $x$-$y$ section from the same location of each sample. It can be seen from Figure 19 that, through only changing $\lambda_x$, the pores are becoming increasingly connected along the $x$-axis while keeping the connection almost unchanged along another direction. This means that the anisotropic conditioning does work, and the pores connection can be modified along one specific direction. In addition, we also test the effects of conditioning of anisotropic correlation length on permeability. We adopt the same samples used in Figure 18 to calculate anisotropic permeability, and plot the results in Figure 20. From the figure, it can be seen that the permeability of generated samples in the $y$- and $z$-axis directions witness an obvious decrease compared to that of training ones, while the permeability of generated samples in the $x$-axis direction exhibits a slight increase. This phenomenon can further validate that the correlation length is positively correlated with permeability statistically. Through the above tests and discussions, we can conclude that the proposed conditional ProGAN can work in both isotropic and anisotropic correlation length conditioning, to provide synthetic samples with modified pore connections.

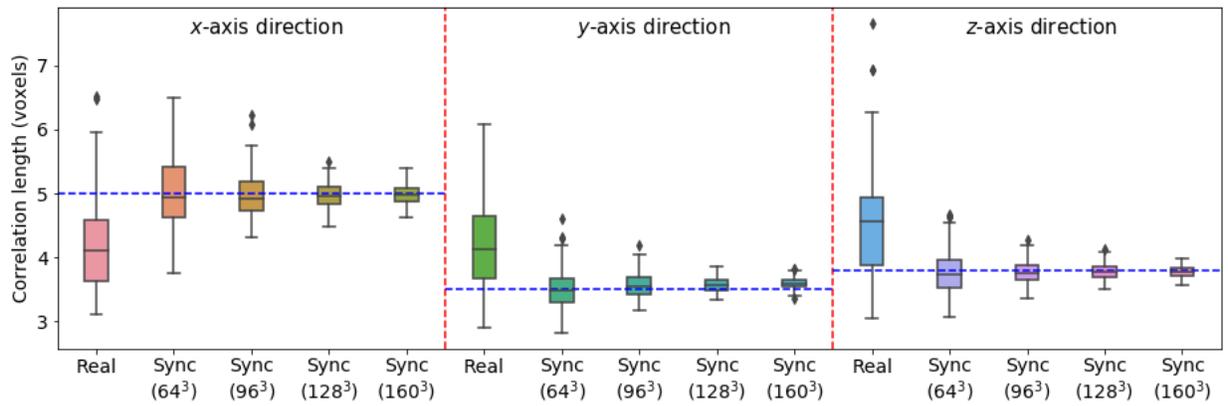

**Figure 18.** Anisotropic correlation lengths (voxels) of the Doddington training samples and generated samples with different sizes. The blue dashed lines represent the respective preset targets for each direction.

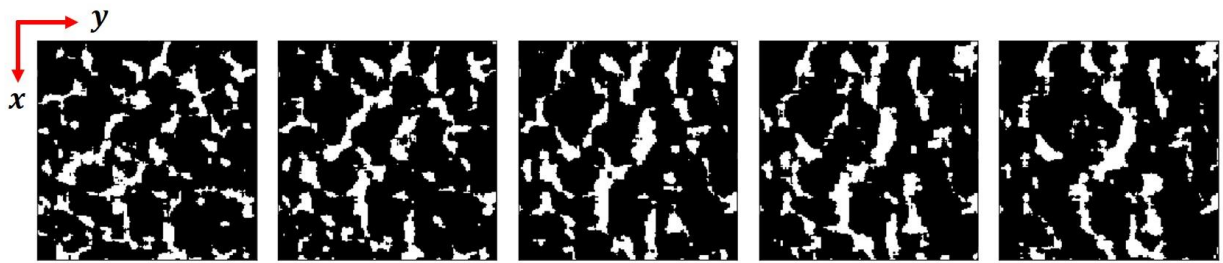

**Figure 19.** $x$-$y$ sections ($160 \times 160$) of generated Doddington samples extracted at a randomly selected but same location, which are produced by one fixed input noise and changing correlation lengths, i.e., increasing $\lambda_x$ from 3.5 voxels to 6.0 voxels with approximately equal intervals (i.e., 3.50, 4.13, 4.75, 5.38, 6.00) and maintaining $\lambda_y = 3.5$ voxels and $\lambda_z = 3.8$ voxels.

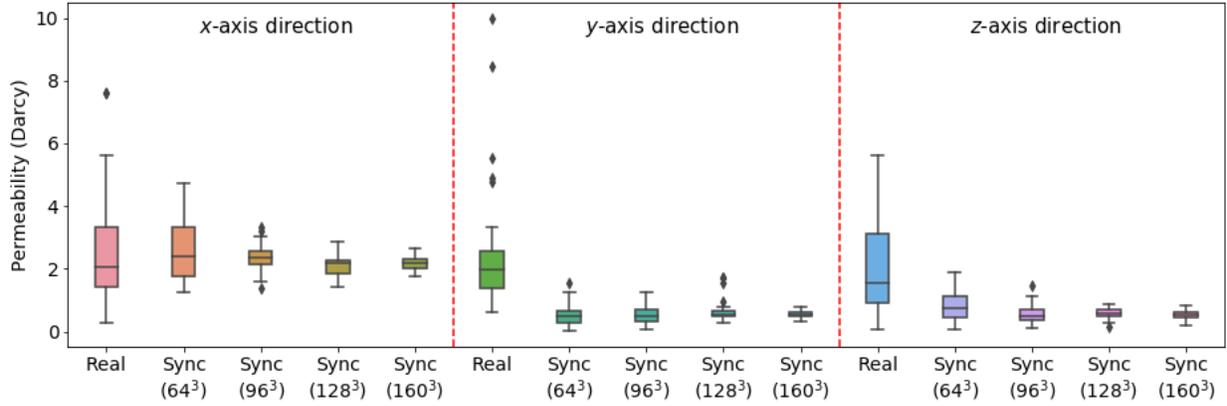

**Figure 20.** Anisotropic permeability of Doddington training samples and generated samples with different sizes. The samples are the same as those used in Figure 18, which adopt different preset correlation lengths as targets in three directions.

## 4. Conclusion

In this work, we extended the framework of digital rock random reconstruction with GANs by introducing a conditioning manner, aiming to produce random samples with user-specified rock types and first two moments. In order to stabilize the training process and naturally learn multiscale features, the conditional ProGAN using WGAN-GP loss was adopted in this work. Through three experiments, it was successfully verified that rock type, porosity, and correlation length could be effectively conditioned on to guide the morphology and geostatistical property of the generated samples without losing visual and geologic realism. Due to the conditioning manner, this study offers a new way to simultaneously incorporate experimental data and user-defined properties when reconstructing digital rocks. Specifically, user-defined properties actually play the role of prior information, which was less utilized in data-driven workflow.

Regarding comparisons with previous representative methods, the commonly used two-point statistics methods can only consider user-defined low-order moments, and subsequent multi-point statistics methods extract high-order information directly from the training images while leaving no interface for users to specify low-order properties. Therefore, the proposed framework in this work actually integrated their advantages, i.e., learning high-order information directly from images using the GANs scheme, while honoring low-order properties through additional conditioning manner. Moreover, the conditioning of rock type may provide an alternative solution to previous isolated GANs training developed for a specific rock type, and consequently save computational cost. Additionally, a potential deficiency exists about rock type conditioning that needs to be mentioned. Since different kinds of rocks have different REV sizes, we refer to their maximum value to design the shape of training samples, aiming to make it satisfy the REV demands of all rock types simultaneously. However, even though the shape may be slightly large for those with small REV size, it will not negatively affect the learning performance of the proposed method.

An alternative research about how to manipulate the generated images is disentangled representation learning, which has been increasingly promoted to improve the interpretability of GANs (Chen et al., 2016; He et al., 2019; Karras et al., 2019) or other generative models, such as variational auto-encoders (Chen et al., 2018; Higgins et al., 2016). The disentanglement of the latent space, including conditional labels, i.e., making them uncorrelated and connected to individual aspects of the generated data, constitutes the key part in such kind of research. Inspired by this concept and in order to clearly demonstrate the effects of each conditional label, we did not carry out the experiment conditioned on the porosity and correlation length together because they are correlated in the binary microstructure (Lu and Zhang, 2002). If the data are continuous, such as the heterogeneous field of geological parameters, it is certain that the first two moments can be conditioned on simultaneously since they are uncorrelated.

## Acknowledgments

This work is partially funded by the Shenzhen Key Laboratory of Natural Gas Hydrates (Grant No. ZDSYS20200421111201738), the SUSTech - Qingdao New Energy Technology Research Institute and the China Postdoctoral Science Foundation (Grant No. 2020M682830). The training and resulting data can be obtained from a public repository by visiting the following URL https://doi.org/10.6084/m9.figshare.14658759.v2.